\documentclass[runningheads]{llncs}
\usepackage{graphicx}

\usepackage{booktabs}
\usepackage{multirow}
\usepackage{amsfonts}
\usepackage{amsmath}
\usepackage{amssymb}
\usepackage{siunitx}
\usepackage{appendix}
\usepackage{hyperref}
\begin{document}
\title{Unsupervised 3D out-of-distribution detection with latent diffusion models}
\titlerunning{3D OOD with latent diffusion models}
%
\author{
Mark S. Graham \inst{1} \and
Walter Hugo Lopez Pinaya \inst{1} \and
Paul Wright \inst{1} \and
Petru-Daniel Tudosiu \inst{1} \and
Yee H. Mah \inst{1,2} \and
James T. Teo \inst{2,3} \and
H. Rolf Jäger \inst{4} \and
David Werring \inst{5} \and
Parashkev Nachev \inst{4} \and
Sebastien Ourselin \inst{1} \and
M. Jorge Cardoso \inst{1}
}

\authorrunning{
M. S. Graham et al.
}

%
\institute{
Department of Biomedical Engineering, School of Biomedical Engineering \& Imaging Sciences,
King’s College London, UK \\\email{mark.graham@kcl.ac.uk}
\and
King's College Hospital NHS Foundation Trust, Denmark Hill, London, UK
\and
Institute of Psychiatry, Psychology \& Neuroscience, King's College London, UK
\and
Institute of Neurology, University College London, UK
\and
Stroke Research Centre, UCL Queen Square Institute of Neurology, London, UK
}
\maketitle              
\begin{abstract}
Methods for out-of-distribution (OOD) detection that scale to 3D data are crucial components of any real-world clinical deep learning system. Classic denoising diffusion probabilistic models (DDPMs) have been recently proposed as a robust way to perform reconstruction-based OOD detection on 2D datasets, but do not trivially scale to 3D data. In this work, we propose to use Latent Diffusion Models (LDMs), which enable the scaling of DDPMs to high-resolution 3D medical data. We validate the proposed approach on near- and far-OOD datasets and compare it to a recently proposed, 3D-enabled approach using Latent Transformer Models (LTMs). Not only does the proposed LDM-based approach achieve statistically significant better performance, it also shows less sensitivity to the underlying latent representation, more favourable memory scaling, and produces better spatial anomaly maps. Code is available at \url{https://github.com/marksgraham/ddpm-ood}.
\end{abstract}
\section{Introduction}
Methods for out-of-distribution (OOD) detection are a crucial component of any machine learning pipeline that is deployed in the real world. They are particularly necessary for pipelines that employ neural networks, which perform well on data drawn from the distribution they were trained on but can produce unexpected results when given OOD data. For medical applications, methods for OOD detection must be able to detect both far-OOD data, such as images of a different organ or modality to the in-distribution data, and near-OOD data, such as in-distribution data corrupted by imaging artefacts. It is also necessary that these methods can operate on high-resolution 3D data. In this work, we focus on methods trained in a fully unsupervised way; without any labels or access to OOD data at train time.

Recently, Latent Transformer Models (LTMs) \cite{esser2021taming} have proven themselves to be effective for anomaly detection and synthesis in medical data \cite{pinaya2022unsupervised,patel2022cross,tudosiu2022morphology}. These two-stage models first use a VQ-VAE \cite{oord2017neural} or VQ-GAN \cite{esser2021taming} to provide a compressed, discrete representation of the imaging data. An autoregressive Transformer \cite{vaswani2017attention} can then be trained on a flattened sequence of this representation. LTMs are particularly valuable in medical data, where the high input size makes training a Transformer on raw pixels infeasible. Recently, these models have been shown to be effective for 3D OOD detection by using the Transformer's likelihood of the compressed sequence to identify both far- and near-OOD samples \cite{graham2022transformer}. These models can also provide spatial anomaly maps that highlight the regions of the image considered to be OOD, particularly valuable for highlighting localised artefacts in near-OOD data.

However, LTMs have some disadvantages. Firstly, likelihood models have well documented weaknesses when used for OOD detection \cite{nalisnick2018deep,choi2018waic,hendrycks2018deep}, caused by focusing on low-level image features \cite{havtorn2021hierarchical,serra2019input}. It can help to measure likelihood in a more abstract representation space, such as that provided by a VQ-VAE \cite{dieleman2020typicality},  but how to train models that provide optimal representations for assessing likelihood is still an open research problem. For example, \cite{graham2022transformer} showed in an ablation study that LTMs fail at OOD when lower levels of VQ-VAE compression are used. Secondly, the memory requirements of Transformers mean that even with high compression rates, the technique cannot scale to very high-resolution medical data, such as a whole-body CT with an image dimension $512^3$. Finally, the spatial anomalies maps produced by LTMs are low resolution, being in the space of the latent representation rather than that of the image itself.

A promising avenue for OOD detection is denoising diffusion probabilistic model (DDPM)-based OOD detection \cite{graham2023denoising}. This approach works by taking the input images and noising them multiple times to different noise levels. The model is used to denoise each of these noised images, which are compared to the input; the key idea is that the model will only successfully denoise in-distribution (ID) data. The method has shown promising results on 2D data \cite{graham2023denoising} but cannot be trivially extended to 3D; as even extending DDPMs to work on high-resolution 2D data is an area of active research. We propose to scale it to 3D volumetric data through the use of Latent Diffusion Models (LDMs). These models, analogous to LTMs, use a first-stage VQ-GAN to compress the input. The DDPM then learns to denoise these compressed representations, which are then decoded and their similarity to the input image is measured directly in the original image space.

The proposed LDM-based OOD detection offers the potential to address the three disadvantages of an LTM-based approach. Firstly, as the method is not likelihood based, it is not necessary that the VQ-GAN provides an ill-defined `good representation'. Rather, the only requirement is that it reconstructs the inputs well, something easy to quantify using reconstruction quality metrics. Secondly, DDPMs have more favourable memory scaling behaviour than Transformers, allowing them to be trained on higher-dimensional representations. Finally, as the comparisons are performed at the native resolution, LDMs can produce high-resolution spatial anomaly maps. We evaluate both the LTM and the proposed LDM model on several far- and near-OOD detection tasks and show that LDMs overcome the three main failings of LTMs: that their performance is less reliant on the quality of the first stage model, that they can be trained on higher dimensional inputs, and that they produce higher resolution anomaly maps.

\section{Methods}
We begin with a brief overview of LDMs and relevant notation before describing how they are used for OOD detection and to estimate spatial anomaly maps.

\subsection{Latent Diffusion Models}
LDMs are trained in two stages. A first stage model, here a VQ-GAN, is trained to compress the input image into a latent representation. A DDPM \cite{ho2020denoising} is trained to learn to sample from the distribution of these latent representations through iterative denoising.

\textbf{VQ-GAN}: The VQ-GAN operates on a 3D input of size $\mathbf{x} \in \mathbb{R}^{H \times W \times D}$ and consists of an encoder $E$ that compresses to a latent space $\mathbf{z} \in \mathbb{R}^{h\times w \times d  \times n}$, where $n$ is the dimension of the latent embedding vector. This representation is quantised by looking up the nearest value of each representation in a codebook containing $K$ elements and replacing the embedding vector of length $d$ with the codebook index, $k$, producing  $\mathbf{z_q} \in \mathbb{R}^{h\times w \times d}$. A decoder $G$ operates on this quantised representation to produce a reconstruction, $\mathbf{\hat{x}} \in \mathbb{R}^{H\times W\times D}$.

In a VQ-VAE \cite{oord2017neural}, $E$, $G$ and the codebook are jointly learnt with a $L_2$ loss on the reconstructions and a codebook loss. The VG-GAN  \cite{esser2021taming} aims to produce higher quality reconstructions by employing a discriminator $D$ and training adversarially, and including a perceptual loss component \cite{zhang2018unreasonable} in addition to the $L_2$ reconstruction loss. Following \cite{tudosiu2020neuromorphologicaly}, we also add a spectral loss component to the reconstruction losses \cite{dhariwal2020jukebox}.

The encoder and decoder are convolutional networks of $l$ levels. There is a simple relationship between the spatial dimension of the latent space, the input, and number of levels: $h, w, d = \frac{H}{2^l},\frac{W}{2^l},\frac{D}{2^l}$, so the latent space is $2^{3l}$ times smaller spatially than the input image, with a $4 \times 2^{3l}$ reduction in memory size when accounting for the conversion from a float to integer representation. In practice, most works use $l=3$ ($512\times$ spatial compression) or $l=4$ ($4096\times$ spatial compression); it is challenging to train a VQ-GAN at higher compression rates.

\textbf{DDPM:} A DDPM is then trained on the latent embedding $\mathbf{z}$ (the de-quantised latent). During training, noise is added to $\mathbf{z}$ according to a timestep $t$ and a fixed Gaussian noise schedule defined by $\beta_t$ to produce noised samples $\mathbf{z}_t$, such that \begin{equation}
    q(\mathbf{z}_t|\mathbf{z}_0) = \mathcal{N}\left(\mathbf{z}_t| \sqrt{\bar{\alpha}_t}\mathbf{z}_0,(1-\bar{\alpha})\mathbf{I} \right)
\end{equation} 
where we use $\mathbf{z}_0$ to refer to the noise-free latent $\mathbf{z}$, we have $0 \leq t \leq T$, and $\alpha_t :=1-\beta_t$ and $\bar{\alpha}_t := \prod_{s=1}^{t}\alpha_s$. We design $\beta_t$ to increase with $t$ such that the latent $\mathbf{z}_T$ is close to an isotropic Gaussian. We seek to train a network that can perform the reverse or denoising process, which can also be written as a Gaussian transition:

\begin{equation}
p_\theta(\mathbf{z}_{t-1}|\mathbf{z}_{t})= \mathcal{N}\left(\mathbf{z}_{t-1}| \boldsymbol{\mu}_\theta(\mathbf{z}_t,t),\mathbf{\Sigma}_\theta(\boldsymbol{z}_t,t)\right)
\end{equation} 

In practice, following \cite{ho2020denoising}, we can train a network $\boldsymbol{\epsilon}_{\theta}(\mathbf{z}_t,t)$ to directly predict the noise used in the forward noising process, $\boldsymbol{\epsilon}$. We can train with a simplified loss 
    $L_\text{simple}(\theta) = \mathbb{E}_{t,\mathbf{z}_0,\boldsymbol{\epsilon}}
    \left[
    \lVert\boldsymbol{
    \epsilon} -\boldsymbol{
    \epsilon}_{\theta}\left( \mathbf{z}_t\right)\rVert^2
    \right]$, and denoise according to 

\begin{equation}
    \mathbf{z}_{t-1}=\frac{1}{\sqrt{\alpha_t}}\left(\mathbf{z}_t-\frac{\beta_t}{\sqrt{1-\bar{\alpha}_t}} \boldsymbol{\epsilon}_\theta\left(\mathbf{z}_t, t\right)\right)+\sigma_t \mathbf{n}
\end{equation}
where $\mathbf{n} \sim \mathcal{N}(\mathbf{0}, \mathbf{I})$.

While in most applications an isotropic Gaussian is drawn and iteratively denoised to draw samples from the model, in this work, we take a latent input $\mathbf{z}_0$ and noise to $\mathbf{z}_t$ for a range of values of $t<T$ and obtain their reconstructions, $\mathbf{\hat{z}}_{0,t} = p_\theta(\mathbf{z}_0|\mathbf{z}_t)$.

\subsection{OOD detection with LDMs}
In \cite{graham2023denoising}, an input image $\mathbf{x}$ that has been noised to a range of $t$-values spanning the range $0 < t < T$ is then denoised to obtain $\mathbf{\hat{x}}_{0,t}$, and we measure the similarity for each reconstruction, $\mathbf{S}(\mathbf{\hat{x}}_{0,t}, \mathbf{x})$. These multiple similarity measures are then combined to produce a single score per input, with a high similarity score suggesting the input is more in-distribution. Typically, reconstruction methods work by reconstruction through some information bottleneck - for an autoencoder, this might be the dimension of the latent space; for a denoising model, this is the amount of noise applied - with the principal that ID images will be successfully reconstructed through the bottleneck, yielding high similarity with the input, and OOD images will not. Prior works have shown the performance becomes dependent on the choice of the bottleneck - too small and even ID inputs are poorly reconstructed, too large and OOD inputs are well reconstructed \cite{lyudchik2016outlier,pimentel2014review,denouden2018improving,zong2018deep}. Reconstructing from multiple $t$-values addresses this problem by considering reconstructions from multiple bottlenecks per image, outperforming prior reconstruction-based methods \cite{graham2023denoising}.

In order to scale to 3D data, we reconstruct an input $\mathbf{x}$ in the latent space of the VQ-GAN, $\mathbf{z}= E\left(\mathbf{x}\right)$. Reconstructions are performed using the PLMS sampler \cite{liu2021pseudo}, which allows for high-quality reconstructions with significantly fewer reconstruction steps. The similarity is measured in the original image space by decoding the reconstructed latents, $\mathbf{S}\left(G\left(\mathbf{\hat{z}}_{0,t}\right),\mathbf{x}\right)$. As recommended by \cite{graham2023denoising}, we measure both the mean-squared error (MSE) and the perceptual similarity \cite{zhang2018unreasonable} for each reconstruction, yielding a total of $2N$ similarity measures for the $N$ reconstructions performed. As the perceptual loss operates on 2D images, we measure it on all slices in the coronal, axial, and sagittal planes and average these values to produce a single value per 3D volume. Each similarity metric is converted into a $z$-score using mean and standard deviation parameters calculated on the validation set, and are then averaged to produce a single score.

\subsection{Spatial anomaly maps}
To highlight spatial anomalies, we aggregate a set of reconstruction error maps. We select reconstructions from $t$-values $=[100,200,300,400]$, calculate the pixel-wise mean absolute error (MAE), $z$-score these MAE maps using the pixel-wise mean and standard deviation from the validation set, and then average to produce a single spatial map per input image.

\section{Experiments}
\subsection{Data} 
We use three datasets to test the ability of our method to flag OOD values in both the near- and far-OOD cases. The \textbf{CROMIS} dataset \cite{cromistrial,wilson2018cerebral} consists of 683 head CT scans and was used as the train and validation set for all models, with a 614/69 split. The \textbf{KCH} dataset consists of 47 head CTs acquired independently from CROMIS, and was used as the in-distribution test set. To produce near-OOD data, a number of corruptions were applied to this dataset, designed to represent a number of acquisition/ data preprocessing errors. These were: addition of Gaussian noise to the images at three levels ($\sigma=0.01,0.1,0.2$), setting the background to values different to the 0 used during training (0.3, 0.6, 1), inverting the image through either of the three imaging planes, removing a chunk of adjacent slices from either the top or centre of the volume, skull-stripping (the models were trained on unstripped images), and setting all pixel values to either $1\%$ or $10\%$ of their true values (imitating an error in intensity scaling during preprocessing). Applying each corruption to each ID image yielded a total of 705 near-OOD images. The \textbf{Decathlon} dataset \cite{antonelli2022medical} comprises a range of 3D imaging volumes that are not head CTs and was used to represent far-OOD data. We selected 22 images from each of the ten classes.  All CT head images were affinely registered to MNI space, resampled to 1\unit{\mm} isotropic, and cropped to a $176\times208\times176$ grid. For the images in the Decathlon dataset, all were resampled to be 1\unit{\mm} isotropic and either cropped or zero-padded depending on size to produce a $176\times216\times176$ grid. All CT images had their intensities clamped between $[-15,100]$ and then rescaled to lie in the range $[0,1]$. All non-CT images were rescaled based on their minimum and maximum values to lie in the $[0,1]$ range.

\subsection{Implementation details}
 All models were implemented in PyTorch v1.13.1 using the MONAI framework v1.1.0 \cite{cardoso2022monai}. Code is available at \url{https://github.com/marksgraham/ddpm-ood}. LTM model code can be found at  \url{https://github.com/marksgraham/transformer-ood}.

\textbf{LDMs:} VQ-GANS were trained with levels $l=2$, 3, or 4 levels with 1 convolutional layer and 3 residual blocks per level, each with 128 channels. Training with $l=3/4$ represents standard practice, training with $l=2$ ($64\times$ spatial compression) was done to simulate a situation with higher-resolution input data. All VQ-GANs had an embedding dim of 64, and the 2, 3, 4 level models have a codebook size of 64, 256, 1024, respectively. Models were trained with a perceptual loss weight of 0.001, an adversarial weight loss of 0.01, and all other losses unweighted. Models were trained with a batch size of 64 for 500 epochs on an A100, using the Adam optimizer \cite{kingma2014adam} with a learning rate of $3\times10^{-4}$ and early stopping if the validation loss did not decrease over 15 epochs. The LDM used a time-conditioned UNet architecture as in \cite{rombach2022high}, with three levels with (128, 256, 256) channels, 1  residual block per level, and attention in the deepest level only. The noise schedule had $T=1000$ steps with a scaled linear noise schedule with $\beta_0=0.0015$ and $\beta_T=0.0195$. Models were trained with a batch size of 112 on an A100 with the Adam optimizer, learning rate $2.5\times10^{-5}$ for 12,000 epochs, with early stopping. During reconstruction, the PLMS scheduler was used with 100 timesteps. Reconstructions were performed from 50 $t$ values spaced evenly over the interval $[0,1000]$.

\textbf{LTM:} The Latent Transformer Models were trained on the same VQ-GAN bases using the procedure described in \cite{graham2022transformer}, using a 22-layer Transformer with dimension 256 in the attention layers and 8 attention heads. The authors in \cite{graham2022transformer} used the Performer architecture \cite{choromanski2020rethinking}, which uses a linear approximation to the attention matrix to reduce memory costs and enable training on larger sequence lengths. Instead, we use the recently introduced memory efficient attention mechanism \cite{rabe2021self} to calculate exact attention with reduced memory costs. This enables us to train a full Transformer on a 3-level VQ-GAN embedding, with a sequence length of $22 \times 27 \times 22=13,068$. Neither the Performer nor the memory-efficient Transformer was able to train on the 2-level embedding, with a sequence length of $44 \times 52 \times 44=100,672$. Models were trained on an A100 with a batch size of 128 using Adam with a learning rate of $10^{-4}$.

\section{Results \& Discussion}
Results and associated statistical tests are shown in Table~\ref{tab:main_results} as AUC scores, with tests for differences in AUC performed using Delong's method \cite{delong1988comparing}. At 4-levels, the LDM and LTM both perform well, albeit with the proposed LDM performing better on certain OOD datasets. LTM performance degrades when trained on a 3-level model, but LDM performance remains high. The 3-level LTM result is in agreement with the findings in \cite{graham2022transformer}. This is likely caused by the previously discussed tendency for likelihood-based models, such as Transformers, to be sensitive to the quality of the underlying representation. For instance, \cite{havtorn2021hierarchical} showed that likelihood-based models can fail unless forced to focus on high-level image features. We posit that at the high compression rates of a 4-level VQ-GAN the representation encodes higher-level features, but at 3-levels the representation can encode lower-level features, making it harder for likelihood-based models to perform well. By contrast, the LDM-based method only requires that the VG-GAN produces reasonable reconstructions. While memory constraints prevented training a 2-level LTM, the more modest requirements on the UNet-based LDM meant it was possible to train. This result has implications for the application of very high-resolution medical data: for instance, a whole-body CT with an image dimension $512^3$ would have a latent dimension $32^3$ even with 4-level compression, too large to train an LTM on but comfortably within the reach of a LDM. The 2-level LDM had reduced performance on two classes that have many pixels with an intensity close to 0 (Hippocampal MR, and Scaling $1\%$). Recent research shows that at higher resolutions, the effective SNR increases if the noise schedule is kept constant \cite{hoogeboom2023simple}. It seems this effect made it possible for the 2-level LDM to reconstruct these two OOD classes with low error for many values of $t$. In future work we will look into scaling the noise schedule with LDM input size.

\begin{table}[t!]
\setlength{\tabcolsep}{6pt}
\centering
\small
\caption{AUC scores for identifying OOD data, with the CT-2 dataset used as the in-distribution test set. Results shown split according to the number of levels in the VQ-GAN. Tests for difference in AUC compare each LTM and LDM models with the same VQ-GAN base, \textbf{bold values} are differences significant with $p<0.001$ and \underline{underlined values} significant with $p<0.05$. Results are shown as N/A for the 2-level LTM as it was not possible to train a Transformer on such a long sequence.}
\label{tab:main_results}
  \begin{tabular}{p{0.3cm}l cc|cc|cc }
  \toprule
& \bfseries Dataset   &\multicolumn{6}{c}{\bfseries Model}\\

& & \multicolumn{2}{c}{2-level} & \multicolumn{2}{c}{3-level}& \multicolumn{2}{c}{4-level}\\
  && LTM & LDM & LTM & LDM & LTM & LDM \\
\midrule
    
\multirow{10}{*}{\rotatebox[origin=c]{90}{Far-OOD}}
&Head MR & N/A  & 72 & 0 & \textbf{100} & 100 & 100 \\
&Colon CT & N/A  & 100 & 100 & 100 & 100 & 100 \\
&Hepatic CT & N/A  & 100 & 100 & 100 & 99.9 & 100 \\
&Hippocampal MR & N/A  & 3.51 & 0 & \textbf{100} & 100 & 100 \\
&Liver CT & N/A  & 100 & 100 & 100 & 99.8 & 100 \\
&Lung CT & N/A  & 100 & 89 & 100 & 100 & 100 \\
&Pancreas CT & N/A  & 100 & 100 & 100 & 99.3 & 100 \\
&Prostate MR & N/A  & 99.9 & 0 & \textbf{100} & 100 & 100 \\
&Spleen CT & N/A  & 100 & 100 & 100 & 99.6 & 100 \\
&Cardiac MR & N/A  & 100 & 90 & 100 & 100 & 100 \\
\midrule
\multirow{14}{*}{\rotatebox[origin=c]{90}{Near-OOD}}
&Noise $\sigma=0.01$ & N/A  & 59.7 & 48.1 & 59.3 & 50.7 & 54.5 \\
&Noise $\sigma=0.1$ & N/A  & 100 & 57.5 & \textbf{100} & 44.7 & \textbf{100} \\
&Noise $\sigma=0.2$ & N/A  & 100 & 88.3 & \underline{100} & 45.6 & \textbf{100} \\
&BG value=0.3 & N/A  & 100 & 100 & 100 & 100 & 100 \\
&BG value=0.6 & N/A  & 100 & 100 & 100 & 100 & 100 \\
&BG value=1.0 & N/A  & 100 & 100 & 100 & 100 & 100 \\
&Flip L-R & N/A  & 53.5 & 49.4 & 61.2 & 51.1 & 58.6 \\
&Flip A-P & N/A  & 100 & 65.6 & \textbf{100} & 90.7 & 100 \\
&Flip I-S & N/A  & 100 & 69.7 & \textbf{100} & 90.5 & 100 \\
&Chunk top & N/A  & 46.1 & 28.6 & \textbf{94.6} & 97.6 & 99.8 \\
&Chunk middle & N/A  & 94.4 & 22 & \textbf{100} & 96.2 & 100 \\
&Skull stripped & N/A  & 98.1 & 0 & \textbf{100} & 100 & 100 \\
&Scaling 1\% & N/A  & 0.317 & 0 & \textbf{100} & 100 & 100 \\
&Scaling 10\% & N/A  & 100 & 0 & \textbf{100} & 100 & 100 \\

\bottomrule
\end{tabular}
\end{table}

\begin{figure}[!t]
\includegraphics[trim={7 55 10 30},clip, width=0.79\textwidth]{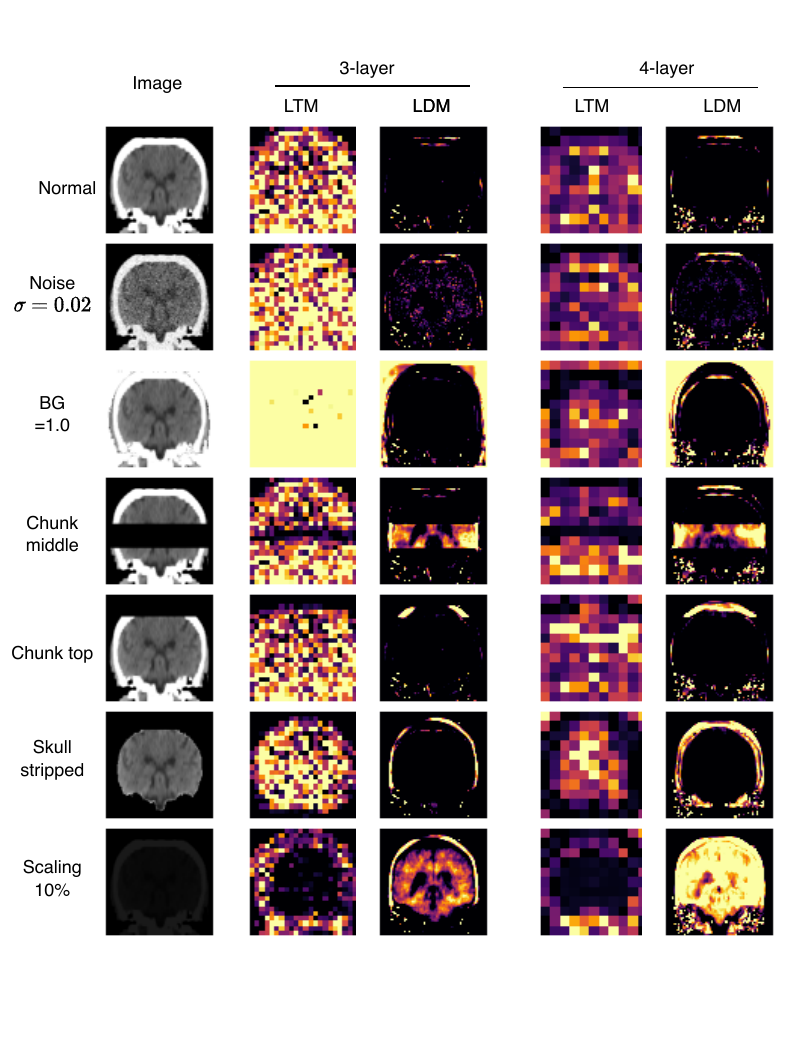}
\centering
\label{fig:anomaly_maps}
\caption{Example anomaly maps for models based on 3- and 4-level VQ-GANs. Maps for each model are shown on the same colour scale, but the scales vary between each model to obtain the best display for each model. Brighter regions are more anomalous.} 
\end{figure}

Anomaly maps are shown in Figure~\ref{fig:anomaly_maps} for near-OOD cases with a spatially localised anomaly. The LDM-based maps are high-resolution, as they are generated in image space, and localise the relevant anomalies. The LTM maps are lower resolution, as they are generated in latent space, but more significantly often fail to localise the relevant anomalies. This is most obvious in anomalies that cause missing signal, such as missing chunks, skulls, or image scaling, which are flagged as low-anomaly regions. This is caused by the tendency of likelihood-based models to view regions with low complexity, such as blank areas, as high-likelihood \cite{serra2019input}. The anomaly is sometimes picked up but not well localised, notable in the `chunk top' example at 4-levels. Here, the transition between brain tissue and the missing chunk is flagged as anomalous rather than the chunk itself.

Memory and time requirements for all models are tabulated in Supplementary \ref{appendix:time_and_memory}. These confirm the LDM's reduced memory use compared to the LTM. All models run in $<30s$, making them feasible in a clinical setting.

\section{Conclusion}
We have introduced Latent Diffusion Models for 3D out-of-distribution detection. Our method outperforms the recently proposed Latent Transformer Model when assessed on both near- and far-OOD data. Moreover, we show LDMs address three key weaknesses of LTMs: their performance is less sensitive to the quality of the latent representation they are trained on, they have more favourable memory scaling that allows them to be trained on higher resolution inputs, and they provide higher resolution and more accurate spatial anomaly maps. Overall, LDMs show tremendous potential as a general-purpose tool for OOD detection on high-resolution 3D medical imaging data. 

\subsubsection{Acknowledgements}
MSG, WHLP, RG, PW, PN, SO, and MJC are supported by the Wellcome Trust (WT213038/Z/18/Z). MJC and SO are also supported by the Wellcome/EPSRC Centre for Medical Engineering (WT203148/Z/16/Z), and the InnovateUK-funded London AI centre for Value-based Healthcare. PTD is supported by the EPSRC (EP/R513064/1). YM is supported by an MRC Clinical Academic Research Partnership grant (MR/T005351/1). PN is also supported by the UCLH NIHR Biomedical Research Centre. Datasets CROMIS and KCH were used with ethics 20/ES/0005.


%
\bibliographystyle{splncs04}
\bibliography{bibliography.bib}

\newpage
\appendix
\title{Supplementary material \\ Unsupervised 3D out-of-distribution detection
with latent diffusion models}
\author{}
\institute{}
\maketitle

\section{Time and memory use}
\label{appendix:time_and_memory}
\begin{table}
\setlength{\tabcolsep}{5pt}
\centering
\small
\caption{Details the memory usage and inference time for a single input. Values include the time/memory required by the VQ-GAN base model, too, which increases for larger numbers of levels, somewhat countering the effect of the Transformer/DDPM operating on lower-resolution data as the levels increase.}
\label{tab:memory_time}
  \begin{tabular}{l cc |cc | cc }
  \toprule
  &\multicolumn{6}{c}{\bfseries Model}\\

&  \multicolumn{2}{c}{2-level} & \multicolumn{2}{c}{3-level}& \multicolumn{2}{c}{4-level}\\
  & LTM & LDM & LTM & LDM & LTM & LDM \\
  \midrule
  Memory/GB & N/A & 1.8& 3.4 &1.8 & 3.4 & 1.5 \\
  Inference time/s & N/A& 28.2 & 2.2 & 14.5 & 0.6 & 12.7\\

\bottomrule
\end{tabular}
\end{table}

\end{document}